\documentclass[11pt]{article}

\usepackage{acl}
\usepackage{times}
\usepackage{latexsym}
\usepackage{booktabs}
\usepackage[T1]{fontenc}
\usepackage[utf8]{inputenc}
\usepackage{inconsolata}
\usepackage{graphicx}
\usepackage{amsmath}
\usepackage[most]{tcolorbox}
\usepackage{CJKutf8} 
\usepackage{geometry}
\usepackage{makecell}
\usepackage{subcaption}
\usepackage{float} 
\usepackage{fontawesome}
\usepackage{enumitem}
\usepackage{amsfonts}
\usepackage[table]{xcolor}
\usepackage{dblfloatfix}
\usepackage{multirow}

\makeatletter
\newif\if@appendixtoc
\@appendixtocfalse
\newcommand{\appendixTOCstart}{\global\@appendixtoctrue}

\let\old@contentsline\contentsline
\renewcommand{\contentsline}[4]{%
  \if@appendixtoc
    \old@contentsline{#1}{#2}{#3}{#4}%
  \fi
}
\makeatother

\title{StreamMeCo: Long-Term Agent Memory Compression for Efficient Streaming Video Understanding}

\author{
{\bfseries Junxi Wang\textsuperscript{1,2},
Te Sun\textsuperscript{1},
Jiayi Zhu\textsuperscript{1},
Junxian Li\textsuperscript{1}}\\[0.1em]
{\bfseries Haowen Xu\textsuperscript{1},
Zichen Wen\textsuperscript{1,3},
Xuming Hu\textsuperscript{4},
Zhiyu Li\textsuperscript{5},
Linfeng Zhang\textsuperscript{1}\thanks{Corresponding author.}}\\[0.1em]
\textsuperscript{1}Shanghai Jiao Tong University,\hspace{0.1cm}
\textsuperscript{2}Fudan University\\[0.1em]
\textsuperscript{3}Shanghai AI Laboratory,\hspace{0.1cm}
\textsuperscript{4}Hong Kong University of Science and Technology\\[0.1em]
\textsuperscript{5}MemTensor (Shanghai) Technology Co., Ltd.\\[0.1em]
\texttt{junxiwang182@gmail.com},\hspace{0.2cm}\texttt{zhanglinfeng@sjtu.edu.cn}
}

\begin{document}
\maketitle

\begin{abstract}

Vision agent memory has shown remarkable effectiveness in streaming video understanding. However, storing such memory for videos incurs substantial memory overhead, leading to high costs in both storage and computation.
To address this issue, we propose {\textbf{StreamMeCo}}, an efficient {\textbf{Stream}} {Agent} {\textbf{Me}mory} {\textbf{Co}mpression} framework. 
Specifically, based on the connectivity of the memory graph, StreamMeCo introduces edge-free minmax sampling for the isolated nodes and an edge-aware weight pruning for connected nodes, evicting the redundant memory nodes while maintaining the accuracy.
In addition, we introduce a time-decay memory retrieval mechanism to further eliminate the performance degradation caused by memory compression. Extensive experiments on three challenging benchmark datasets (M3-Bench-robot, M3-Bench-web and Video-MME-Long) demonstrate that under \textbf{70\%} memory graph compression, StreamMeCo achieves a \textbf{1.87$\times$} speedup in memory retrieval while delivering an average accuracy improvement of \textbf{1.0\%}. \emph{Our code is available at https://github.com/Celina-love-sweet/StreamMeCo}.

\end{abstract}

\section{Introduction}

With the rapidly growing demands of application scenarios such as live streaming \cite{r23}, real-time surveillance \cite{r24}, and autonomous driving \cite{r25}, research on streaming video understanding has become increasingly important. Unlike offline video understanding \cite{r1, r2, r3, r4}, where models can access the complete video content and user questions prior to inference, streaming video understanding \cite{r10, r13, r15, r17, r18, r21} can only rely on the limited information observed before the user's question arrives, making how to efficiently process continuously incoming visual information a critical challenge.
Current research on streaming video understanding primarily focuses on achieving long-video understanding by carefully handling the vision tokens and KV cache \cite{r8, r12}. However, for extremely long videos, these methods still suffer from the loss of critical visual information and struggle to maintain long-term consistency of entities such as human identities.  
\begin{figure}[t] 
  \centering  
  \includegraphics[width=\linewidth]{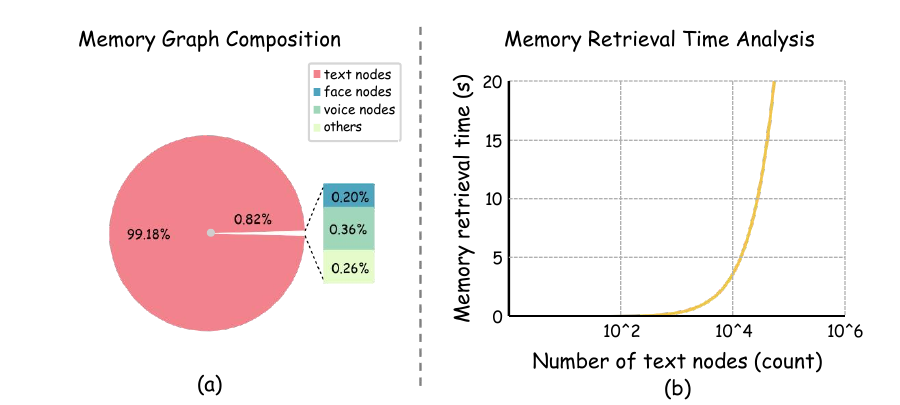}  
  \vspace{-15pt}
  \caption{An example from M3-Bench-robot. (a) Composition of the memory graph. (b) Analysis of memory retrieval time with respect to the number of text nodes, indicating that the \emph{increment of memory nodes makes the retrieval time not acceptable.}}
  \label{Fig1}
  \vspace{-15pt} 
\end{figure}

To tackle this challenge, agent memory-based methods are introduced to organize the information of videos as memory in plaintext. For example,
M3-Agent \cite{r22} leverages agent memory mechanism to model streaming video as a memory graph, representing entities with face and voice nodes and abstracting rich visual content into text nodes, thereby preserving information integrity and long-term entity consistency. Nevertheless, as the memory graph scales up, storage and retrieval efficiency become major bottlenecks. As shown in Figure~\ref{Fig1}, the increasing number of text nodes leads to a rapidly increasing memory retrieval latency, which hinders the real-time answering of questions.

To address this, we propose \textit{\textbf{StreamMeCo}}, an efficient training-free Long-Term \textit{\textbf{Stream}} \textit{Agent} \textit{\textbf{Me}mory} \textit{\textbf{Co}mpression} framework, achieving the first memory compression of industrial-level streaming video agent and supporting efficient memory graph retrieval. Based on the connectivity of the memory graph, we design a dual-branch memory compression method. For isolated text nodes, we use the \textit{\textbf{Edge-free Minmax sampling (EMsampling)}} module for compression. First, we cluster the text nodes into different groups, then for each cluster, we first select the node closest to the cluster center and add it to the selected set, then calculate the minimum distance between each unselected node and the selected set, choosing the farthest point. This process continues until the number of selected nodes meets the predefined requirement. For connected text nodes, we use the \textit{\textbf{Edge-aware Weighting pruning (EWpruning)}} module. By calculating the weight edge matrix between text nodes and face nodes or voice nodes, we determine the entity importance of each text node. We then combine the embedding similarity of the text nodes to select the most dissimilar text nodes, and ultimately determine the retained text nodes based on a comprehensive fusion score.

Additionally, to reduce the performance degradation caused by compressing the memory graph, inspired by the human memory mechanism, we design a \textit{\textbf{Time-decay Memory Retrieval (TMR)}} mechanism for efficiently retrieving the memory graph. We calculate the total similarity between text nodes in the memory graph and the model's query for each time segment. Based on the total similarity of each time segment, we dynamically allocate the number of text nodes to different segments. Furthermore, to simulate the memory decay mechanism of the human brain, we assign nonlinear time-varying decay to the text nodes of each time segment, thereby adaptively balancing long-term memory and recent critical information during the retrieval stage, and guaranteeing the retrieval of more memories from recent segments.

In summary, the main contributions of this paper are as follows:
\begin{itemize}[leftmargin=1em, itemsep=0pt, topsep=0pt]
    \item We propose \textit{{edge-free minmax sampling}} module and \textit{{edge-aware weighting pruning }} module, which efficiently compress the memory graph based on its structural properties.
    \item We introduce a novel \textit{time-decay memory retrieval } mechanism, which has been experimentally proven to significantly outperform previous retrieval frameworks.
    \item We demonstrate the effectiveness of \textit{StreamMeCo} through extensive experiments. On three high-difficulty benchmark datasets, under approximately \textbf{70\%} memory graph compression, the proposed method achieves a \textbf{1.87$\times$} speedup in memory retrieval while delivering an average accuracy improvement of \textbf{1.0\%}.
\end{itemize}

\section{Related Work}

\subsection{Streaming Video Understanding}

Streaming video understanding enables a model to continuously process incoming frames, and upon receiving a user question \(Q_t\) at time \(t\), generate responses based on the accumulated content \(C_{1:t}\). Existing methods fall into two main paradigms: visual token compression and KV cache compression. The former selects informative tokens either in pixel space or after feature encoding \cite{r5, r6, r7, r12, r14, r16, r19}. While the latter maintains a short-term sliding window and a long-term memory stack at the LLM level to preserve information across temporal scales \cite{r8, r9, r11, r13, r20}. Recently, M3-Agent \cite{r22} departs from these designs by integrating long-term agent memory into streaming video understanding, showing strong potential for modeling long-range temporal dependencies and maintaining entity consistency.

\subsection{Long-Term Agent Memory}

Long-term agent memory is designed to provide LLM-based agents with persistent memory beyond a single context window. Depending on the type of input it handles, it can be categorized into unimodal and multimodal forms. Unimodal memory \cite{r26, r27, r28, r29} typically maintains consistency across multi-turn reasoning by constructing structured representations, latent embeddings, or summaries. In contrast, multimodal memory \cite{r30, r31, r32, r33} integrates visual, linguistic, and other signals into a unified space, updating them to support stable long-term reasoning and decision-making. Building on this foundation, M3-Agent is the first to introduce long-term agent memory into streaming video scenarios, enabling application in real-time settings.

\subsection{LLMs and Agent Systems}

With the development of LLMs and MLLMs, they have demonstrated strong performance across a variety of core tasks, including retrieval and question answering \cite{r47, r48}, audio and video understanding \cite{r49, r50, r51}, temporal modeling \cite{r52}, and complex reasoning \cite{r53, r54, r55}. Meanwhile, a range of research directions aimed at enhancing their capabilities have also rapidly emerged, such as VLA modeling \cite{r56}, unified multimodal modeling \cite{r57}, inference acceleration \cite{r58, r59}, and chain-of-thought reasoning optimization \cite{r60, r61, r62}. In addition, LLMs and MLLMs have been widely applied in agent frameworks, covering tool use and policy optimization \cite{r63, r64, r65}, multi-agent collaborative planning \cite{r66, r69}, and agentic reinforcement learning \cite{r67}, significantly improving their ability to handle complex real-world scenarios. In the domain of agent memory, a variety of memory frameworks built upon LLMs and MLLMs, such as Mem0 \cite{r45}, MemOS \cite{r46}, and M3-Agent \cite{r22}, have shown great potential in long-term information storage and retrieval, further highlighting the central role of large models in intelligent agent systems.

\begin{figure}[t] 
  \centering  
  \includegraphics[width=\linewidth]{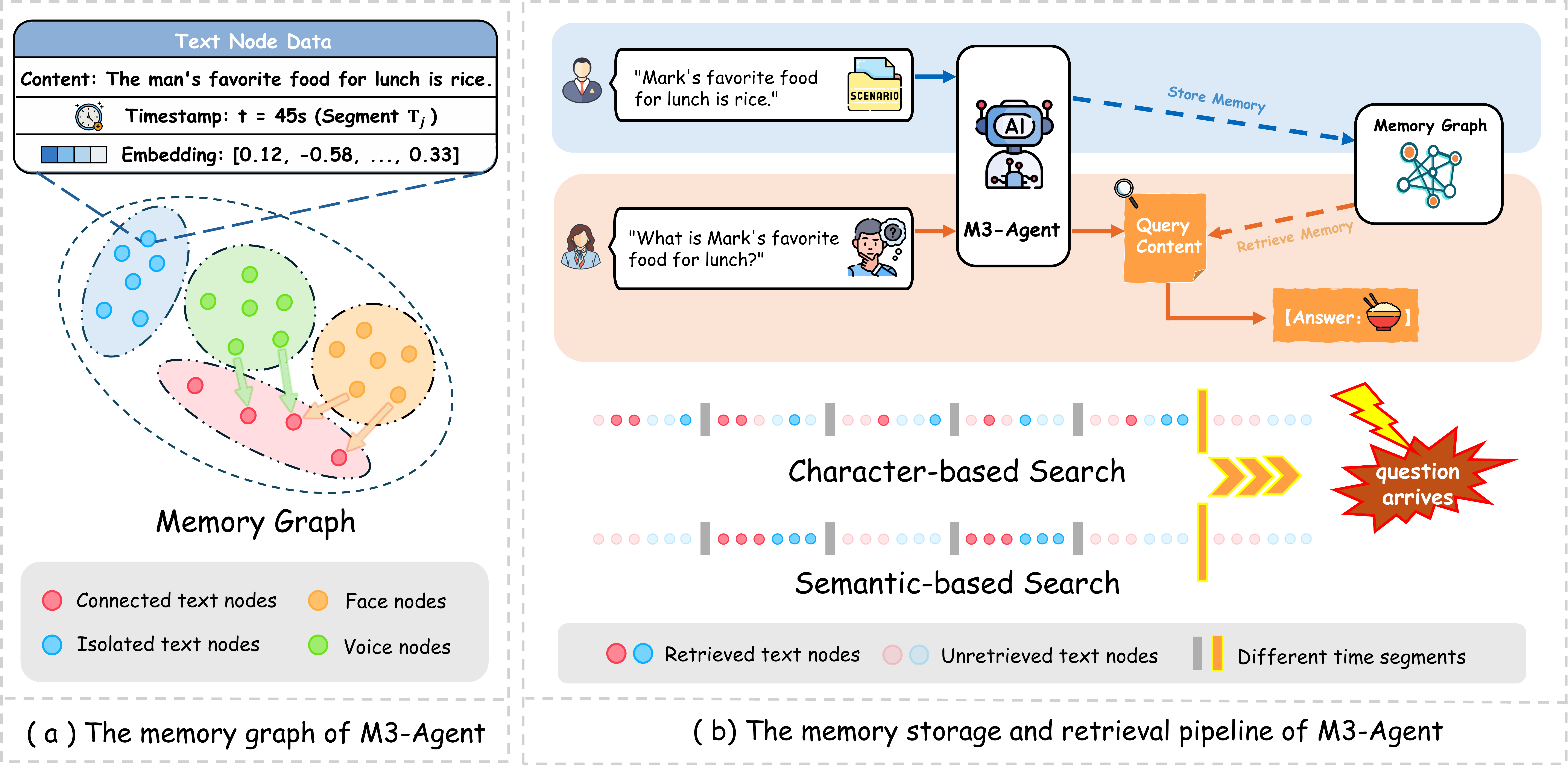}  
  \vspace{-15pt}
  \caption{The overview of M3-Agent. (a) The memory graph of M3-Agent. (b) The memory storage and retrieval pipeline of M3-Agent, illustrating two different retrieval strategies.}
  \label{Fig2}
  \vspace{-15pt} 
\end{figure}

\section{Preliminaries}

Graph-based agent memory has demonstrated great potential across a variety of research areas. Our method exhibits strong generality and can be readily adapted to different graph-based agent memory frameworks, while this paper focuses on its application in streaming video understanding scenarios. Next, we provide a detailed introduction to M3-Agent \cite{r22}.

\subsection{Memory Graph of M3-Agent}

As shown in Figure~\ref{Fig2}(a), the memory graph includes face nodes, voice nodes, text nodes, and other relevant information such as edge weights. Each text node contains content, embedding vectors, and a timestamp of the content. Among these nodes, only some voice nodes are connected to text nodes, and some face nodes are connected to text nodes, while there are no internal connections within face nodes, voice nodes, or text nodes, nor between voice and face nodes. Based on the graph's connectivity, the content of isolated text nodes is similar to: "\textit{The man's favorite food for lunch is rice.}" with no specific entity references. In contrast, the content of connected text nodes is like: "\textit{Mark's favorite food for lunch is rice.}" which explicitly includes entity information.

\subsection{Retrieval of Memory Graph}
\label{Section3.2}

As shown in Figure~\ref{Fig2}(b), after a question is input into M3-Agent, the model performs reasoning and outputs the query content to be retrieved. Based on this content, the model retrieves from the memory graph, which is primarily divided into two approaches: \textit{\textbf{character-based}} and \textit{\textbf{semantic-based}}. If the retrieval content contains characters like “\textit{character id}”, the model returns the \textit{k} most similar content based on the similarity between the embedding of the query content and the content embedding of the text nodes, as the retrieved memory. If there is no “\textit{character id}” identifier, semantic retrieval is performed. In this case, the model segments all text nodes based on their timestamps, selecting the most similar node as the representative for each time segment. Finally, the two most similar text nodes are selected, and all text node contents at the timestamps of these two nodes are returned as the retrieved memory.

\section{Methodology}

\begin{figure*}[t]
  \centering  
  \includegraphics[width=\textwidth]{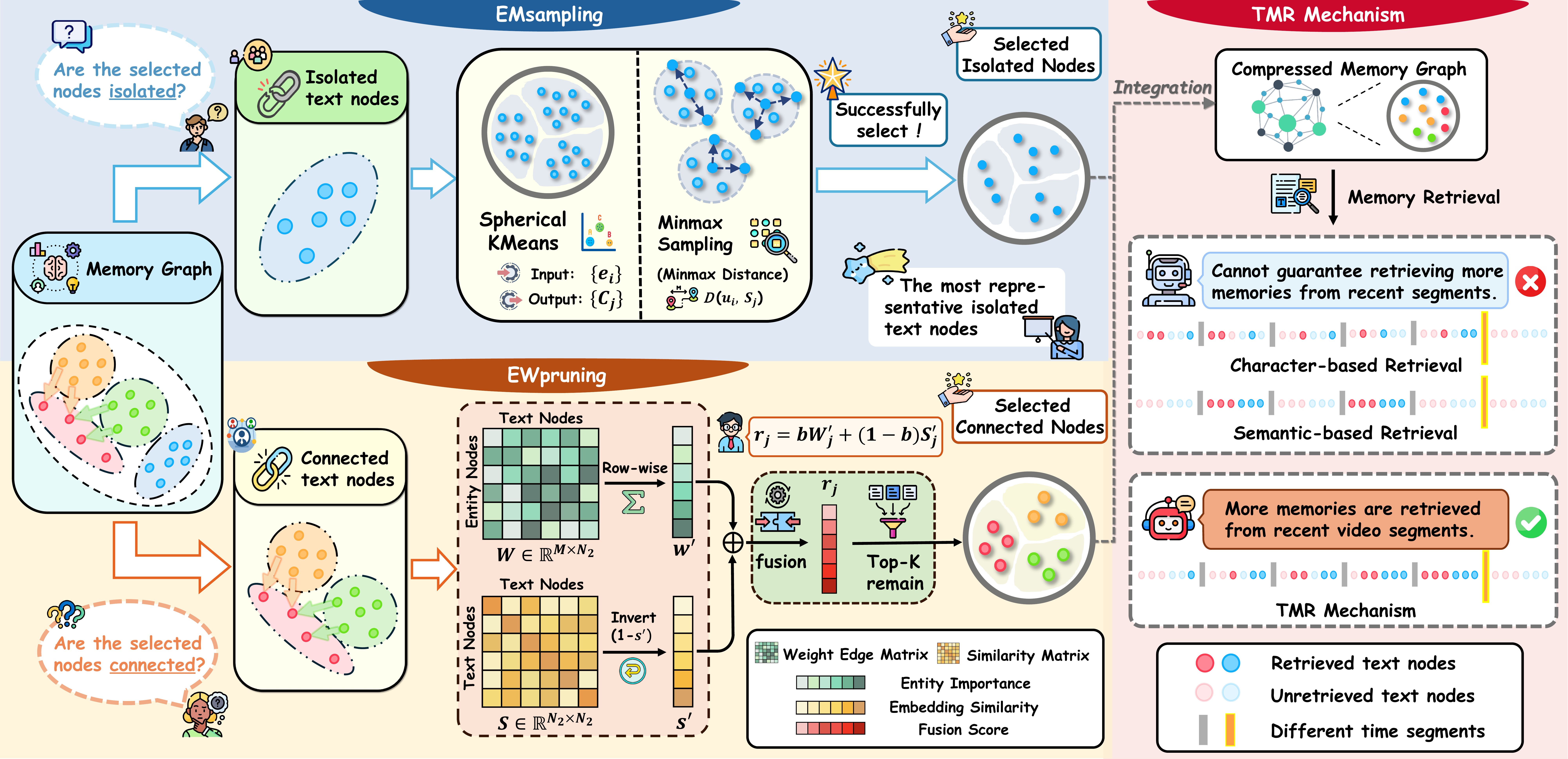}  
  \vspace{-15pt} 
  \caption{The overview of StreamMeCo. Efficient memory graph compression is achieved via the EMsampling and EWpruning modules, while the TMR mechanism prioritizes retrieving memories from more recent segments, enabling fast and accurate memory retrieval.}
  \label{Fig3}
  \vspace{-15pt} 
\end{figure*}

\subsection{Overview}

The overview of StreamMeCo is illustrated in Figure~\ref{Fig3}. The framework consists of two components, achieving efficient compression of memory graph while supporting fast and accurate retrieval.

\subsection{Text Memory Compression}

Based on the connectivity of the memory graph, we categorize text nodes into isolated nodes and connected nodes, and accordingly propose a dual-branch memory compression method. Different compression strategies are applied to these two types of text nodes.

\paragraph{Edge-free Minmax Sampling.}

For isolated text nodes, we have carefully designed the EMsampling module to select the most representative text nodes. Specifically, given \( N_1 \) text nodes, we use the \textit{Spherical KMeans Algorithm} \cite{r41} to cluster them into \( N_1 \times a \) clusters based on the embeddings of the text node contents, as shown in the following formula:

\begin{equation}
\{C_j\} = \textit{KMeans}(\{e_i\}, a),
\label{eq1}
\end{equation}
where \( e_i \) (\( i = 1, 2, \dots, N_1 \)) represents the embedding vector of the text node contents, and clustering ratio \(a\) is used to control the number of clusters, while \( C_j \) (\( j = 1, 2, \dots, N_1 \times a \)) represents the \( j \)-th cluster.

Then, we assume the retention ratio of isolated text nodes is \( \alpha \), and distribute them evenly across each cluster. The number of text nodes retained in the \( j \)-th cluster is \( |C_j| \times \alpha \), where \( |C_j| \) represents the number of text nodes in the \( j \)-th cluster.

Finally, we adopt an effective \textit{minmax sampling} strategy to select the text nodes to be retained from each cluster. Specifically, for the \( j \)-th cluster, we first select the text node closest to the cluster center and add it to the selected node set \( S_j \). Then, we compute the distance from each unselected text node in the unselected node set \( U_j \) to the selected node set \( S_j \), and select the node with the largest distance to add to \( S_j \). The distance from each unselected node to the selected node set \( S_j \) is defined as the minimum distance to any node in the selected set, as shown in the following formula:

\begin{equation}
D(u_i, S_j) = \min_{s_k \in S_j} \| e_{u_i} - e_{s_k} \|,
\label{eq2}
\end{equation}
where \( u_i \in U_j \) is an unselected node, and \( s_k \in S_j \) is a selected node, while \( e_{u_i} \) and \( e_{s_k} \) represent the embedding vectors of the unselected and selected nodes, respectively. 

We select unselected node \( u_{i^*} \) with the farthest distance and add it to the selected node set \( S_j \):

\begin{equation}
u_{i^*} = \arg\max_{u_i \in U_j} D(u_i, S_j).
\label{eq3}
\end{equation}
Until the number of selected nodes reaches \( |C_j| \times \alpha \), the number of retained text nodes.

\paragraph{Edge-aware Weighting Pruning.}

For connected text nodes, we carefully design the EWpruning module, thereby constructing a more compact yet informative memory representation to enable efficient node pruning. Specifically, given \( N_2 \) text nodes that are connected to face and voice nodes via edges, we construct an entity-based weight edge matrix \( W \in \mathbb{R}^{M \times N_2} \), where \( M \) denotes the total number of face and voice nodes. Each entry \( w_{ij} \) represents the edge weight between the \( i \)-th face or voice node and the \( j \)-th text node, reflecting the importance of the \( j \)-th text node to the \( i \)-th face or voice node.

For the weight matrix \( W \), the row-wise summation is computed as \( W' = \sum_{i=1}^{M} w_{ij} \), where \( W' \in \mathbb{R}^{1 \times N_2} \) represents the importance scores of the \( N_2 \) connected text nodes with respect to the set of face and voice entity nodes. Subsequently, we normalize \( W' \) to the range \([0, 1]\) to obtain the final \textit{entity importance} for each text node.

Next, for the \( N_2 \) text nodes, we compute their pairwise similarities to obtain a similarity matrix \( S \in \mathbb{R}^{N_2 \times N_2} \), where \( s_{ij} \) denotes the embedding similarity between the \( i \)-th and \( j \)-th text nodes. We then perform a row-wise summation on the similarity matrix, computed as \( S' = \sum_{i=1}^{N_2} s_{ij} \), where \( S' \in \mathbb{R}^{1 \times N_2} \) represents the total similarity of each text node to all other text nodes. Since we aim to select the text nodes with higher diversity, we normalize \( S' \) to the range \([0, 1]\) and then invert it as \( 1 - S' \), resulting in the \textit{embedding similarity} for each text node relative to the connected text node set. For brevity, we continue to denote the processed entity importance and embedding similarity as \( W' \) and \( S' \).

We assume the retention ratio for connected text nodes is \( \beta \), meaning that we need to keep a total of \( N_2 \times \beta \) text nodes. For the \( j \)-th connected text node, we compute its normalized entity importance \( W'_j \) and its embedding similarity \( S'_j \) through a weighted combination. The final fusion score is computed as:

\begin{equation}
r_j = b W'_j + (1 - b) S'_j,
\label{eq6}
\end{equation}
where the balancing coefficient \( b \) controls the trade-off between entity importance and embedding similarity. 

Finally, we retain the top \( N_2 \times \beta \) text nodes with the highest scores and prune the others.

\subsection{Time-decay Memory Retrieval}

To reduce the performance degradation caused by compressing the memory graph, inspired by the human memory mechanism, we propose a TMR framework. Specifically, given the query content generated by the model in Section~\ref{Section3.2}, we first compute the embedding similarity between the query and the content embeddings of all text nodes. According to their timestamps, these text nodes are grouped into a series of chronologically ordered time segments \( T_j \) (\( j = 1, 2, \dots, t \)). At time step \( t \), when the system receives a query, the relevance score of each time segment is computed by aggregating the similarity scores of all text nodes within that segment as \( E_j = \sum_{i=1}^{|T_j|} s_{ij} \), where \( s_{ij} \) denotes the embedding similarity between the query and the \( i \)-th text node in \( T_j \), \( E_j \) denotes the aggregated similarity within the time segment \( T_j \), and \( s_{ij} \) denotes the embedding similarity between the query content and the \( i \)-th text node in time segment \( T_j \), and \( |T_j| \) represents the number of text nodes in segment \( T_j \). 

Next, to simulate human memory decay, we first compute the average similarity of each time segment as \( \bar{E}_j = \frac{E_j}{|T_j|} \). Based on this averaged relevance score, we then apply a temporal decay function to model the fading of older memories:

\begin{equation}
E_j' = \bar{E}_j \cdot e^{-\lambda (t - j)},
\label{eq9}
\end{equation}
where \( j = 1, 2, \dots, t \) indicates the temporal position of the segment, and the memory decay coefficient \( \lambda \) controls the rate of memory decay.

To be consistent with the number of retrieved text nodes in Section~\ref{Section3.2}, we assume that the model needs to retrieve a total of \textit{k} text nodes as the final memory output. Accordingly, we compute the number of nodes selected from each time segment as:

\begin{equation}
\textit{Num}_j = \frac{E_j'}{\sum_{i=1}^{t} E_i'} \times \textit{k},
\label{eq10}
\end{equation}
where \( \textit{Num}_j \) denotes the number of nodes selected from the \( j \)-th segment.

Finally, within each time segment, we select the top-ranked text nodes according to their similarity to the query content until reaching \( \textit{Num}_j \) nodes, which constitute the final retrieved memory.

\section{Experiments}

\subsection{Experiments Setup}

\paragraph{Datasets.}

We select three challenging open-source benchmark datasets, M3-Bench-robot, M3-Bench-web \cite{r22} and Video-MME-Long \cite{r42}, to evaluate the effectiveness of StreamMeCo. M3-Bench-robot is a streaming video dataset, while the others are offline video datasets. Consistent with the original M3-Agent setup, we use GPT-4o to assess the answer quality and employ text-embedding-3-large to encode the query content generated by M3-Agent to supporting subsequent memory graph retrieval. For more details, please refer to the Appendix~\ref{A.1}.

\paragraph{Baselines.}

Our baselines include closed-source MLLMs: Gemini-1.5-Pro \cite{r35}, GPT-4o \cite{r34}, Gemini-Agent and Gemini-GPT4o-Hybrid \cite{r22}. open-source MLLMs: Qwen2.5-VL-7B-Instruct \cite{r37}, Qwen2.5-Omni-7B \cite{r36}. As well as streaming video models: MovieChat \cite{r39}, MA-LMM \cite{r40},  VideoChat-Online \cite{r16}, Flash-VStream \cite{r18}, TimeChat-Online \cite{r12}, StreamForest \cite{r7}, and M3-Agent \cite{r22}. For more introductions about the baselines, please refer to the Appendix~\ref{A.2}.

\paragraph{Implementation Details.}

All experiments are conducted on two NVIDIA A100 (80G) GPUs. To reduce error, each experiment is executed three times and the average result is reported. In our setup, the clustering ratio  \( a \) is set to 0.05, the balancing coefficient \( b \) is set to 0.1, and the memory decay coefficient \(\lambda\) is set to 0.1. For the M3-Bench-robot and M3-Bench-web, we use the memory graphs provided by the M3-Agent paper, for Video-MME-Long, we replace the Gemini-1.5-Pro API with Gemini-2.5-Pro, while keeping all other settings consistent with M3-Agent. We apply compression only to the memory graph produced by M3-Agent, while keeping all other components identical to the original model.

\begin{table*}[t]
\fontsize{8}{13}\selectfont
\setlength{\tabcolsep}{4.1pt}
\centering
\begin{tabular}{lcccccc|cccccc|c|c}
  \toprule[1.5pt]
  \textbf{Dataset}
       & \multicolumn{6}{c}{\textbf{M3-Bench-robot}}
       & \multicolumn{6}{c}{\textbf{M3-Bench-web}}
       & \multirow{2}{*}[-0.8ex]{\shortstack[c]{\textbf{Video-}\\\textbf{MME-Long}}}
       & \multirow{2}{*}[-0.8ex]{\textbf{Avg.}} \\
  \cmidrule(lr){2-7}\cmidrule(lr){8-13}
  \textbf{Model}
       & ME & MH & CM & PU & GK & \textbf{All}
       & ME & MH & CM & PU & GK & \textbf{All}
       &  \\
  \midrule

  \rowcolor{gray!15}
  & \multicolumn{12}{c}{\emph{Closed-source MLLMs}} & \multicolumn{2}{c}{~} \\
  Gemini-1.5-Pro            &6.5&7.5&8.0&9.7&7.6&8.0&18.0&17.9&23.8&23.1&28.7&23.2&38.0&23.1 \\
  GPT-4o                    &9.3&9.0&8.4&10.2&7.3&8.5&21.3&21.9&30.9&27.1&39.6&28.7&38.8&25.3 \\
  Gemini-Agent              &15.8&17.1&15.3&20.0&15.5&16.9&29.3&20.9&33.8&34.6&45.0&34.1&55.1&35.4 \\
  Gemini-GPT4o-Hybrid       &21.3&25.5&22.7&28.8&23.1&24.0&35.9&26.2&37.6&43.8&52.2&41.2&56.5&40.6 \\
  \midrule

  \rowcolor{gray!15}
  & \multicolumn{12}{c}{\emph{Open-source MLLMs}} & \multicolumn{2}{c}{~} \\
  Qwen2.5-VL-7b-Instruct    &2.9&3.8&3.6&4.6&3.4&3.4&11.9&10.5&13.4&14.0&20.9&14.9&46.9&21.7 \\
  Qwen2.5-Omni-7b           &2.1&1.4&1.5&1.5&2.1&2.0&8.9&8.8&13.7&10.8&14.1&11.3&42.2&18.5 \\
  \midrule

  \rowcolor{gray!15}
  & \multicolumn{12}{c}{\emph{Streaming Video Models}} & \multicolumn{2}{c}{~} \\
  MovieChat {\footnotesize\textcolor{gray}{[CVPR24]}} 
                            &13.3&9.8&12.2&15.7&7.0&11.2&12.2&6.6&12.5&17.4&11.1&12.6&19.4&14.4 \\
  MA-LMM {\footnotesize\textcolor{gray}{[CVPR24]}}
                            &25.6&23.4&22.7&39.1&14.4&24.4&26.8&10.5&22.4&39.3&15.8&24.3&17.3&22.0 \\
  VideoChat-Online {\footnotesize\textcolor{gray}{[CVPR25]}}
                            &31.7&24.7&30.5&43.1&17.1&29.9&34.3&18.9&34.1&48.3&23.1&32.7&45.9&36.2 \\
  Flash-VStream {\footnotesize\textcolor{gray}{[ICCV25]}}
                            &21.6&19.4&19.3&24.3&14.1&19.4&24.5&10.3&24.6&32.5&20.2&23.6&25.0&22.7 \\
  TimeChat-Online {\footnotesize\textcolor{gray}{[MM25]}}
                            &35.6&29.4&31.5&44.3&22.3&33.6&38.8&22.8&38.4&51.6&28.0&36.6&49.2&39.8 \\
  StreamForest {\footnotesize\textcolor{gray}{[NIPS25]}}
                            &33.4&29.4&31.3&44.3&19.0&32.1&37.9&22.3&39.9&51.7&27.6&36.5&51.9&40.2 \\
  \rowcolor{blue!3}
  M3-Agent {\footnotesize\textcolor{gray}{[ICLR26]}} 
                            &30.9&29.4&29.6&41.1&22.6&30.3&44.9&25.6&44.8&58.6&53.7&47.9&\textbf{56.0}&44.7 \\
  \rowcolor{blue!3}
  \hspace{15pt}+Random ({\color{green!40!black}$\downarrow 30\%$})
                            &28.9&29.4&27.7&40.5&18.0&28.5&42.4&24.7&40.1&54.8&50.8&44.6&54.2&42.4 \\

  \rowcolor{blue!3}
  \hspace{15pt}+Clustering ({\color{green!40!black}$\downarrow 30\%$})
                            &30.0&32.9&30.0&42.5&20.8&29.6&41.2&26.3&41.7&55.9&50.9&45.3&54.6&43.2 \\
  \rowcolor{blue!3}
  \hspace{15pt}+DART ({\color{green!40!black}$\downarrow 30\%$})
                            &29.5&24.7&25.8&40.1&21.7&29.1&40.4&26.0&41.3&55.6&51.2&45.4&54.8&43.1 \\

  \rowcolor{blue!3}
  \hspace{15pt}+TimeChat-Memory ({\color{green!40!black}$\downarrow 30\%$})
                            &31.8&36.5&31.7&42.2&18.7&30.2&39.8&27.1&38.9&56.6&46.7&44.7&55.0&43.3 \\

  \rowcolor{blue!3}
  \hspace{15pt}+MemoryLLM ({\color{green!40!black}$\downarrow 30\%$})
                            &31.4&34.1&30.9&41.4&18.0&28.9&38.4&24.9&36.6&56.5&45.4&44.5&53.9&42.8 \\

  \rowcolor{blue!3}
  \hspace{15pt}+StreamMeCo ({\color{green!40!black}$\downarrow 30\%$})
                            &32.9&30.6&30.9&42.5&19.6&30.7&41.3&26.0&44.3&58.3&50.5&47.0&54.8&44.2 \\

  \rowcolor{blue!3}
  \hspace{15pt}+StreamMeCo ({\color{green!40!black}$\downarrow 50\%$})
                            &32.3&28.2&29.8&41.4&21.1&30.6&39.7&25.2&38.9&58.4&47.3&44.7&54.4&43.2 \\

  \rowcolor{blue!3}
  \hspace{15pt}+StreamMeCo ({\color{green!40!black}$\downarrow 70\%$})
                            &29.1&30.6&26.9&40.3&19.9&28.4&37.0&21.7&36.1&53.0&46.3&41.9&53.3&41.2 \\
             
  \rowcolor{blue!3}
  \hspace{15pt}+StreamMeCo+TMR ({\color{green!40!black}$\downarrow 30\%$})
                            &35.7&32.9&35.5&44.0&27.5&\textbf{34.6}&46.4&32.2&46.5&61.8&55.8&\textbf{50.7}&\textbf{55.2}&\textbf{46.8} \\
  \rowcolor{blue!3}
  \hspace{15pt}+StreamMeCo+TMR ({\color{green!40!black}$\downarrow 50\%$})
                            &34.8&34.1&32.4&43.1&24.5&\textbf{33.4}&45.4&31.3&45.0&59.2&52.7&\textbf{48.9}&\textbf{56.6}&\textbf{46.3} \\
  \rowcolor{blue!3}
  \hspace{15pt}+StreamMeCo+TMR ({\color{green!40!black}$\downarrow 70\%$}) 
                            &34.9&31.8&32.1&42.7&24.8&\textbf{34.2}&43.8&28.9&43.9&57.9&52.9&\textbf{47.9}&54.9&\textbf{45.7} \\
  \bottomrule[1.5pt]
\end{tabular}
\caption{Performance comparison of different models on M3-Bench-robot, M3-Bench-web and Video-MME-Long. The best three results for the streaming video models are highlighted in \textbf{bold black font}.}
\label{table1}
\vspace{-12pt} 
\end{table*}

\subsection{Main Results}

To comprehensively evaluate the performance of our method, we conduct comparisons with 13 competitive baselines. For more details about the baselines, please refer to the Appendix~\ref{A.3}. As shown in Table~\ref{table1}, the overall performance of \textbf{StreamMeCo} significantly surpasses that of existing closed-source and open-source MLLMs, as well as other streaming video models, demonstrating the effectiveness of the agent memory mechanism for streaming video understanding. Even with \textbf{70\%} of text nodes compressed, StreamMeCo and TMR mechanism still achieves an average accuracy improvement of \textbf{1.0\%} across all datasets compared to the uncompressed M3-Agent.

In addition, we have constructed five memory graph compression methods and compared them with StreamMeCo, with the compression ratio controlled at 30\%. In the random method, we randomly selected nodes from two types of text nodes for compression. For the clustering method, we used the KMeans clustering algorithm to evaluate the differences between our method and traditional clustering approaches. Specifically, we divided the two types of text nodes into a series of smaller clusters, each of which retains the text node closest to the cluster center. For the DART method, we were inspired by the DART \cite{r43}. Specifically, for the two types of text nodes, we first randomly selected 2\% of the pivot nodes, then calculated the total similarity of the remaining nodes with these pivot nodes, prioritizing the compression of nodes with higher similarity to the pivot nodes, until the target compression ratio was reached. Finally, we also draw inspiration from the streaming video understanding model TimeChat-Online \cite{r12} and the LLM memory method MemoryLLM \cite{r68}, which we denote as TimeChat-Memory and MemoryLLM, respectively. For TimeChat-Memory, based on the chronological order of Agent Memory, we compute the similarity between two adjacent memory nodes. If the similarity exceeds 0.7, the earlier memory node is removed; otherwise, both are retained. Meanwhile, following the original design, the memory buffer adopts a first-in-first-out (FIFO) strategy, once the target compression ratio is exceeded, earlier memories are preferentially discarded. For MemoryLLM, a fixed-capacity memory bank is maintained, where memories are continuously written in chronological order. When the memory bank reaches its capacity, each newly incoming memory is compared with the existing ones, and the most similar old memory is removed to make room for the new memory. As shown in the Table~\ref{table1}, StreamMeCo significantly outperforms these methods in terms of performance.

For the M3-Bench-robot dataset, all three compression ratios lead to varying degrees of performance improvement. In contrast, the performance gains on the M3-Bench-web and Video-MME-Long datasets are relatively limited. This is because M3-Bench-robot consists of real-world filming captured from a first-person robotic perspective, involving daily environments and human–robot interactions, which naturally contain a higher level of redundancy. In comparison, both M3-Bench-web and Video-MME-Long are sourced from YouTube, where videos typically exhibit stronger narrative structures or scripted production, resulting in more concise content and relatively less redundancy, leaving limited room for effective compression.

\subsection{Ablation Study}

Due to the increased overlap between the nodes compressed by StreamMeCo and those retained through random compression at higher text-node compression ratios, the performance differences between modules become less distinguishable. To highlight the contribution of each component in our compression, we conduct the components ablation analysis under a 30\% compression ratios.

\paragraph{Compression Components Ablation Analysis.}

\begin{table}[t]
\fontsize{9.25}{13}\selectfont
\centering
\begin{tabular}{ccc|c|c}
\toprule[1.5pt]
\multicolumn{3}{c|}{\textbf{Components}} 
& \textbf{M3-Bench-Robot} 
& \textbf{M3-Bench-Web} \\
\hline
\textbf{A} & \textbf{B} & \textbf{C} 
& Accuracy 
& Accuracy \\
\hline

\rowcolor{gray!15}
-- & -- & -- & 30.3 (100.0\%) & 47.9 (100.0\%) \\

\hline
 &  &  & 28.5 (94.1\%)  & 44.6 (93.1\%)  \\
\checkmark &  &  & 30.3 (100.0\%) & 45.7 (95.4\%) \\
 & \checkmark & \checkmark & 30.1 (99.3\%) & 46.1 (96.2\%) \\
 \checkmark & \checkmark &  & 30.2 (99.7\%) & 46.1 (96.2\%) \\
\checkmark &  & \checkmark & 30.5 (100.7\%) & 46.6 (97.3\%) \\
\hline
\rowcolor{blue!3}
\checkmark & \checkmark & \checkmark & \textbf{30.7 (101.3\%)} & \textbf{47.0 (98.1\%)} \\
\bottomrule[1.5pt]
\end{tabular}

\caption{
Ablation study of the compression components. The first row reports the accuracy of the M3-Agent without compression.
\textbf{A}: EMsampling; 
\textbf{B}: entity importance; 
\textbf{C}: embedding similarity.
}
\vspace{-12pt}
\label{table2}
\end{table}

This component analysis focuses solely on the compression method and does not involve the TMR mechanism. As shown in Table~\ref{table2}, we evaluate the performance of EMsampling (A), entity importance (B), and embedding similarity (C) under different combinations. The results indicate that enabling any single component leads to varying degrees of performance improvement. Specifically, the EMsampling module effectively selects the most representative text nodes from the isolated-node subset, while for connected text nodes, our method jointly considers both entity importance and embedding similarity, yielding a more compact yet informative memory representation. When all three components are used together, the model achieves its best performance on both benchmark datasets, reaching accuracies of \textbf{30.7\%} and \textbf{47.0\%}, corresponding to \textbf{101.3\%} and \textbf{98.1\%} of the original M3-Agent, respectively. In contrast, random compression achieves only 28.5\% and 44.6\% accuracy on the two benchmark datasets, resulting in substantial performance loss.

We observe that compression causes greater performance degradation on M3-Bench-web than on M3-Bench-robot, further supporting our earlier hypothesis that the text nodes in M3-Bench-robot contain higher redundancy.

\paragraph{TMR Impact Analysis.}

\begin{figure}[t]  
  \centering  
  \includegraphics[width=\linewidth]{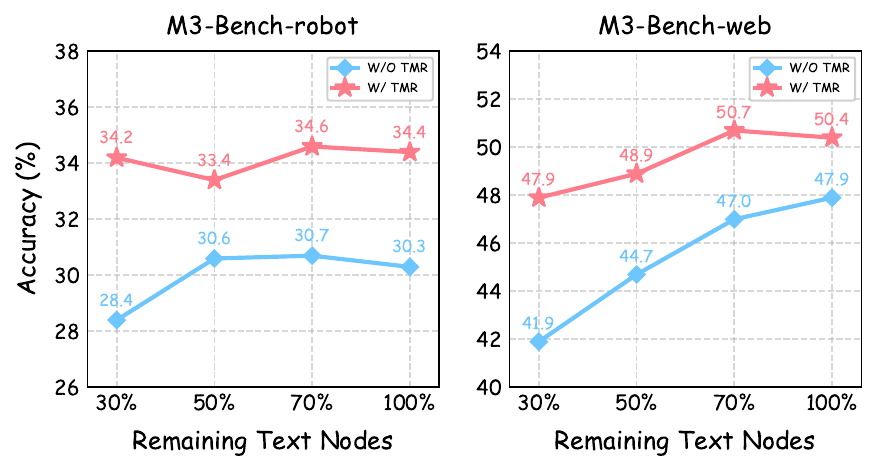}  
  \vspace{-17pt} 
  \caption{Impact of the TMR Mechanism on M3-Bench-robot and M3-Bench-web at different compression ratios.}
  \label{Fig4} 
  \vspace{-15pt} 
\end{figure}

As shown in Figure~\ref{Fig4}, when the TMR mechanism is not applied, the model's performance drops significantly on both benchmark datasets once the compression ratio exceeds 50\%. In contrast, after introducing TMR, the model accuracy improves by \textbf{3.3\%} over the original M3-Agent even without any compression. Compressing 30\% of the text nodes further enhances the performance. More importantly, under a 70\% compression setting, the model does not suffer from noticeable performance degradation and performs comparably to or even better than the uncompressed M3-Agent. These observations demonstrate that TMR mechanism can significantly reduce the accuracy loss caused by memory graph compression and effectively retrieve memories from more recent segments for reasoning.

\paragraph{Time Efficiency Analysis.}

\begin{figure*}[t]
  \centering  
  \includegraphics[width=\textwidth]{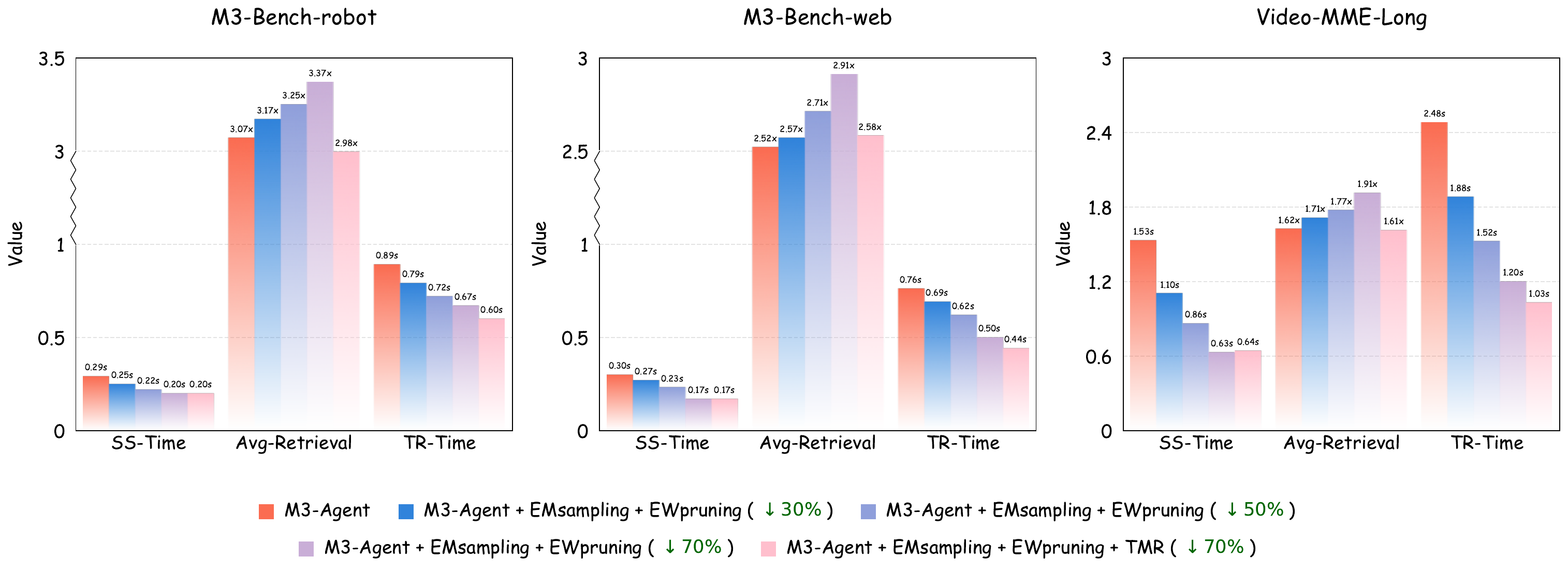}  
  \vspace{-17pt} 
  \caption{Time Efficiency Analysis on M3-Bench-Robot, M3-Bench-Web and Video-MME-Long. \textbf{SS-Time:} Time to compute and sort the similarity between the query embedding and each text node's content embedding; \textbf{Avg-Retrieval:} Average number of memory graph queries per question; \textbf{TR-Time:} Average total time spent querying the memory graph for each question after obtaining the model's query embedding.}
  \label{Fig5}
  \vspace{-12pt} 
\end{figure*}

In this section, we analyze the impact of the EMsampling module, EWpruning module, and TMR mechanism on the time efficiency of M3-Agent, with a focus on the memory retrieval time after converting the model query into embeddings. As shown in Figure~\ref{Fig5}, when only applying the EMsampling and EWpruning modules for compression, SS-Time decreases as the compression ratio increases, while Avg-Retrieval increases correspondingly (M3-Agent triggers up to five memory retrievals per question by default). We attribute this phenomenon to the reduced memory capacity after compression, which limits the amount of effective information the model can retrieve, lowers the confidence in the retrieval results, and requires more retrieval rounds to finalize the answer.

In contrast, with the introduction of the TMR mechanism, Avg-Retrieval significantly decreases, as TMR mechanism prioritizes more reliable and higher-confidence memory entries, enabling the model to obtain sufficient information with fewer retrieval iterations, thus effectively reducing the overall processing time.

When compressing \textbf{70\%} of the memory, our method achieves an average \textbf{1.87$\times$} speedup in memory retrieval time across all datasets.

\paragraph{Effect of Cluster Ratio \textit{a}.}

\begin{figure}[t]  
  \centering  
  \includegraphics[width=\linewidth]{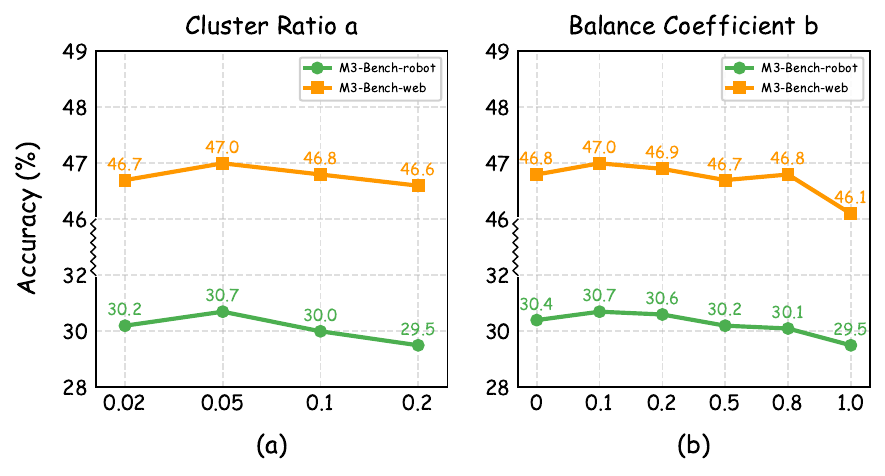}  
  \vspace{-15pt}
  \caption{Impact of cluster ratio \textit{a} and balance boefficient \textit{b} on M3-Bench-robot and M3-Bench-web.}
  \label{Fig6} 
  \vspace{-15pt} 
\end{figure}

As shown in Figure~\ref{Fig6}(a), the model performs best on both datasets when the cluster ratio \textit{\textbf{a = 0.05}}. When \textit{a} is smaller, the number of clusters decreases, and each cluster contains more samples, which makes it more likely that text nodes with large semantic differences are assigned to the same cluster, thereby affecting the preservation of semantic distinctions. In contrast, when \textit{a} is larger, the number of clusters increases, and each cluster contains fewer samples, making it difficult to form stable and representative semantic structures, ultimately degrading the performance of the compressed model.

\paragraph{Effect of Balance Coefficient \textit{b}.}

As shown in Figure~\ref{Fig6}(b), the model achieves the best performance on both datasets when the parameter balance coefficient \textit{\textbf{b = 0.1}}. When \textit{b} is further decreased or increased, the performance drops to varying degrees. This observation indicates that an excessively small balance coefficient prevents the model from effectively leveraging entity-importance information, whereas an excessively large balance coefficient causes the model to overly rely on entity importance, thereby ignoring embedding similarity information and ultimately degrading the overall performance.

\paragraph{On Other Decay Methods}

We first compare linear decay with the exponential decay used in our paper (evaluated on the M3-Bench-Robot dataset). The results are shown in Table~\ref{table3}, where the parameter denotes the decay coefficient. From the table, exponential decay shows a generally monotonic decreasing trend, while linear decay exhibits a rise-then-fall pattern. This observation aligns with intuition: exponential decay decreases more rapidly and better matches the human memory forgetting curve, achieving stronger performance under smaller decay coefficients. We also note that linear decay could theoretically approximate exponential decay, but would require more delicate parameter tuning. Therefore, exponential decay is more practical and robust in real-world scenarios.

\begin{table}[t]
\fontsize{9.5}{15}\selectfont
  \centering
  \begin{tabular}{ccc}
    \toprule[1.5pt]
    \textbf{Decay Coefficient} & \textbf{Exponential} & \textbf{Linear} \\
    \midrule
    0.1 & 34.6 & 33.5 \\
    0.5 & 32.1 & 33.7 \\
    1.0 & 32.9 & 32.4 \\
    2.0 & 32.0 & 33.5 \\
    \bottomrule[1.5pt]
  \end{tabular}
  \caption{Comparison between diffenent decay methods on M3-Bench-Robot.}
  \label{table3}
  \vspace{-12pt}
\end{table}

In addition, we experimented with a piecewise decay strategy: starting from the time segment corresponding to the query, the memory strength is halved every five temporal segments. This strategy achieves an accuracy of 33.8\% on M3-Bench-Robot, which is still lower than exponential decay.

In addition, for more experiments and theoretical analysis, please refer to the appendix.

\section{Conclusion}

We propose \textbf{StreamMeCo}, the first memory compression framework specifically designed for industrial-scale streaming video agent. It achieves efficient compression and accurate retrieval of memory graphs through the integration of our EMsampling and EWpruning modules, along with the TMR mechanism. Experimental results show that even with a \textbf{70\%} reduction in memory graph size, StreamMeCo achieves a \textbf{1.87$\times$} speedup in memory retrieval, while delivering an average accuracy improvement of \textbf{1.0\%}.

\section*{Acknowledge}

This project was supported by New Generation Artificial Intelligence-National Science and Technology Major Project.

\section*{Limitations}

Due to the frequent need to call the Gemini-2.5-Pro and text-embedding-3-large APIs during memory graph generation, as well as the substantial time required to generate the memory graphs, and since M3-Agent currently only constructs and provides memory graphs for M3-Bench-robot and M3-Bench-web, we were limited by budget and time constraints and were only able to test the Video-MME-Long dataset additionally. We hope that future work can test more benchmark datasets to more comprehensively validate the performance of our method.

\bibliography{custom}

\clearpage
\appendix
\addtocontents{toc}{\protect\appendixTOCstart}
\renewcommand{\contentsname}{Appendix}

\begingroup
\setcounter{tocdepth}{3}
\makeatletter

\let\oldl@subsection\l@subsection
\renewcommand{\l@subsection}[2]{%
  \vspace{0.6ex}
  \oldl@subsection{#1}{#2}%
}

\let\oldl@subsubsection\l@subsubsection
\renewcommand{\l@subsubsection}[2]{%
  \vspace{0.6ex}
  \oldl@subsubsection{#1}{#2}%
}

\makeatother
\tableofcontents
\endgroup

\section{Other Experimental Details}

\subsection{Dataset}

\begin{figure}[t] 
  \centering  
  \includegraphics[width=\linewidth]{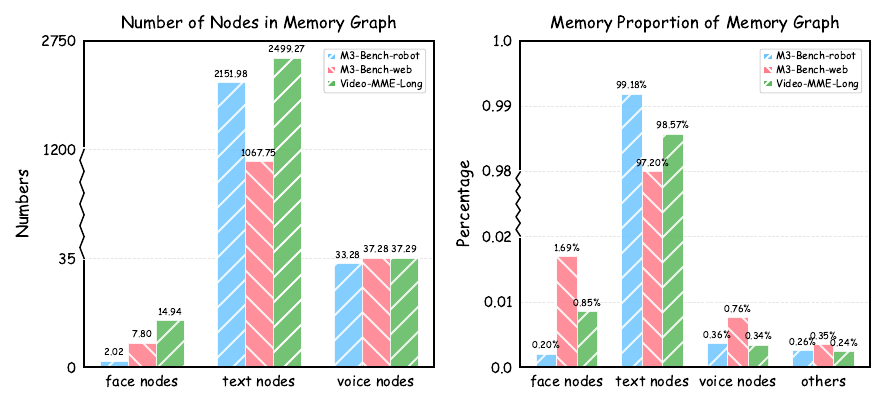}  
  \vspace{-15pt}
  \caption{Numbers of different types of nodes and their memory proportion in the memory graphs of the three benchmark datasets.}
  \label{Fig8}
  \vspace{-15pt} 
\end{figure}

The detailed statistics of the three benchmark datasets are provided in Table~\ref{table4}, and the detailed information of the memory graph is shown in Figure~\ref{Fig8}. Next, we provide a detailed description of each dataset.

\label{A.1}

\begin{table*}[t]
\fontsize{9.5}{15}\selectfont
  \centering
  \begin{tabular}{cccccc}
    \toprule[1.5pt]
    \textbf{Dataset} & \textbf{Avg. Duration(s)} & \textbf{\#Videos} & \textbf{\#QA Pairs}  & \textbf{Avg. Text Nodes} & \textbf{\#Source}\\
    \midrule
    \textbf{M3-Bench-robot} & 2039.9 & 100 & 1276  & 2151.98 & Real-world filming \\
    \textbf{M3-Bench-web}   & 1630.7 & 920 & 3214  & 1067.75 & YouTube\\
    \textbf{Video-MME-Long}   & 2466.7 & 300 & 900  & 2499.27 & YouTube\\
    \bottomrule[1.5pt]
  \end{tabular}
  \caption{The statistical information of M3-Bench-robot, M3-Bench-web and Video-MME-Long.}
  \label{table4}
  \vspace{-12pt}
\end{table*}

\paragraph{M3-Bench-robot.}
M3-Bench-robot contains 100 real-world videos recorded from a robot's first-person perspective, covering seven everyday environments, such as living rooms, kitchens, bedrooms, study rooms, offices, meeting rooms, and gyms. Each video depicts interactions between the robot and two to four different humans, where the robot must build and reason based on these interactions. The dataset includes five reasoning types: ME (multi-evidence reasoning), MH (multi-hop reasoning), CM (cross-modal reasoning), PU (person understanding), and GK (general knowledge extraction). M3-Bench-robot aims to evaluate how robots use long-term memory to reason, particularly focusing on complex spatial relationships, object tracking, and dynamic human interactions.

\paragraph{M3-Bench-web.}
M3-Bench-web contains 920 videos sourced from YouTube, covering 46 different video types such as documentaries, street interactions, variety shows, travel, food, and sports events. The videos provide a wide range of content, designed to challenge the agent's ability to reason in complex and dynamic environments. Like M3-Bench-robot, M3-Bench-web also includes five reasoning types, assessing how the agent integrates cross-modal information, understands relationships between people, and extracts general knowledge. This dataset is ideal for evaluating the agent's memory reasoning capabilities in diverse tasks and complex scenarios, especially for cross-modal reasoning and memory retrieval in varied contexts.

\paragraph{Video-MME-Long.}
Video-MME-Long is the long-video subset of the Video-MME benchmark, consisting of 300 real-world long videos and 900 multiple-choice questions. The videos have an average duration of approximately 41 minutes and may extend up to 1 hour, covering 6 major visual domains and 30 fine-grained categories. The dataset provides multimodal inputs including video frames, audio, and subtitles, while the questions emphasize cross-temporal event association and complex spatiotemporal reasoning. This makes Video-MME-Long a highly challenging benchmark for evaluating multimodal large language models in long-term temporal understanding and multimodal information integration.

\subsection{Baselines}
\label{A.2}

\subsubsection{Closed-source MLLMs}

\paragraph{Gemini-1.5-Pro.} Gemini 1.5 Pro is an advanced multimodal model capable of handling long-contexts of up to 10 million tokens. It is based on a sparse mixture-of-experts (MoE) Transformer architecture, efficiently processing various data types such as text, video, and audio, and performs exceptionally well across multiple benchmarks.

\paragraph{GPT-4o.} GPT-4o is an advanced multimodal model capable of processing a variety of input types, including text, images, and videos. It excels in real-time response, efficiently handling inputs and generating multimodal outputs. The system is particularly optimized for visual understanding, enabling fast and accurate interactions in multimodal tasks. This design allows GPT-4o to perform robustly across diverse challenging multimodal benchmarks.

\subsubsection{Open-source MLLMs}

\paragraph{Qwen2.5-VL.} Qwen2.5-VL is a powerful multimodal vision-language model designed for understanding and reasoning across various input types, including images, videos, and text. It excels in tasks such as precise object localization, document parsing, and long-video comprehension, with breakthroughs in video understanding and object localization. Qwen2.5-VL also supports efficient reasoning across diverse domains, handling complex input data and generating accurate multimodal outputs.

\paragraph{Qwen2.5-Omni.} Qwen2.5-Omni is a unified multimodal model capable of processing various inputs such as text, images, audio, and video, while simultaneously generating text and natural speech responses in a streaming manner. It utilizes an innovative Thinker–Talker architecture, enabling real-time speech responses, and excels in tasks like audio understanding and video reasoning, making it suitable for real-time multimodal interaction scenarios.

\subsubsection{Streaming Video Models}

\paragraph{MovieChat.} MovieChat is a video understanding system that integrates video models and LLM. It uses a sliding window to extract frame-level features and stores them in hybrid memory, with the LLM performing QA based on this memory. Supporting long videos over 10K frames, the system effectively addresses computational complexity and memory overhead, enhancing overall long video comprehension.

\paragraph{MA-LMM.} MA-LMM is a memory-augmented large multimodal model designed for long-term video understanding. It processes video frames sequentially and stores information in a long-term memory bank, avoiding the context length and GPU memory limitations faced by traditional models when handling long videos. The model efficiently captures temporal information in videos and achieves excellent performance in tasks such as video question answering and video captioning.

\paragraph{VideoChat-Online.} VideoChat-Online is an efficient video LLM designed for streaming video understanding, utilizing a Pyramid Memory Bank (PMB) architecture to balance spatial and temporal details. The model is optimized through an offline-to-online learning paradigm, combined with an interleaved dialogue format for video data, enabling real-time video stream processing and outstanding performance across multiple benchmarks. It outperforms many existing models in online video tasks and demonstrates excellent cross-platform capabilities.

\paragraph{Flash-VStream.} Flash-VStream is a real-time video stream understanding model that employs the STAR (Spatial-Temporal-Abstract-Retrieved) memory mechanism, enabling efficient processing of extremely long video streams and real-time user query responses. By compressing visual information and dynamically updating memory, the model significantly reduces inference latency and GPU memory consumption, supporting online video question answering. Flash-VStream outperforms existing methods in multiple benchmarks, especially in streaming video scenarios.

\paragraph{TimeChat-Online.} TimeChat-Online is an advanced online VideoLLM designed for real-time video interaction. Its core innovation lies in the Differential Token Drop (DTD) module, which reduces visual redundancy in streaming videos by selectively preserving significant temporal changes while discarding static content between frames. This approach helps achieve an 82.8\% reduction in video tokens, maintaining high performance in real-time video question answering (QA). Additionally, it integrates proactive responding capabilities, triggering answers based on future visual cues from video scene transitions.

\paragraph{StreamForest.} StreamForest is an online video understanding model specifically designed for streaming video scenarios. Its core innovation lies in the Persistent Event Memory Forest, which dynamically organizes video frames into event-level tree structures, ensuring effective storage and processing of long-term visual information. The model also introduces a Fine-grained Spatiotemporal Window, enhancing real-time scene perception. It performs exceptionally well across multiple streaming video understanding benchmarks, particularly in real-time video question answering and autonomous driving applications.

\paragraph{M3-Agent.} M3-Agent is the first industrial-grade streaming video agent, integrated with a long-term Agent Memory mechanism, capable of processing video and audio inputs in real-time. It generates episodic and semantic memories to gradually build world knowledge and completes tasks through multi-turn reasoning.

\begin{figure}[t]  
  \centering  
  \includegraphics[width=\linewidth]{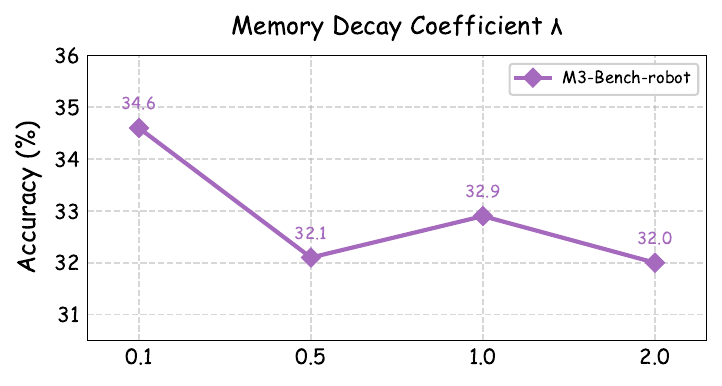}  
  \vspace{-12.5pt}
  \caption{Impact of memory decay coefficient \(\lambda\) on M3-Bench-robot.}
  \label{Fig7} 
  \vspace{-15pt} 
\end{figure}

\subsection{Baselines setup}
\label{A.3}

\subsection{Closed-source MLLMs}
\label{A.3.1}
M3-Agent consists of two models, memorization and control, and is trained using reinforcement learning. Following the approach in the M3-Agent paper, we use the closed-source MLLMs Gemini-1.5-Pro and GPT-4o to replace the memorization model, generating descriptions for video clips and storing them as long-term memory. Subsequently, through retrieval-augmented generation (RAG) ~\cite{r38}, we use GPT-4o to perform memory retrieval as a replacement for the control model.

Following the approach in the M3-Agent paper, for Gemini-Agent, we use Gemini-1.5-Pro to replace the memorization and control models in M3-Agent. For the generated memory graph, we use M3-Agent’s search function to replace RAG. Similarly, for Gemini-GPT4o-Hybrid, we use Gemini-1.5-Pro to generate memory as the memory model, use GPT-4o for control, and still use M3-Agent’s search function to replace RAG.

\subsubsection{Open-source MLLMs}
Following the approach in the M3-Agent paper, we use the open-source MLLMs Qwen2.5-VL and Qwen2.5-Omni to replace the closed-source MLLMs Gemini-1.5-Pro and GPT-4o in Section~\ref{A.3.1}, to ensure a fair comparison, with the rest remaining consistent as before.

\subsection{Streaming Video Models}
For the streaming video models, we use the pretrained models and default parameter settings from the official papers.

\section{Additional Ablation Study}
\label{B}

\subsection{Effect of Memory Decay Coefficient \textit{\textbf{\(\lambda\)}}.}

As shown in Figure~\ref{Fig7}, the model achieves its best performance when the memory decay coefficient is set to memory decay coefficient \textit{\textbf{\(\lambda\) = 0.1}}. As \(\lambda\) increases further, the performance drops to varying degrees. This indicates that an excessively large decay coefficient leads to an overly strong temporal decay effect, causing the model to diminish the importance of historical memory too early. Consequently, the model fails to effectively leverage long-term contextual information, ultimately degrading its reasoning performance.

\subsection{TMR Impact on Video-MME-Long}

\begin{figure}[t]  
  \centering  
  \includegraphics[width=0.63\linewidth]{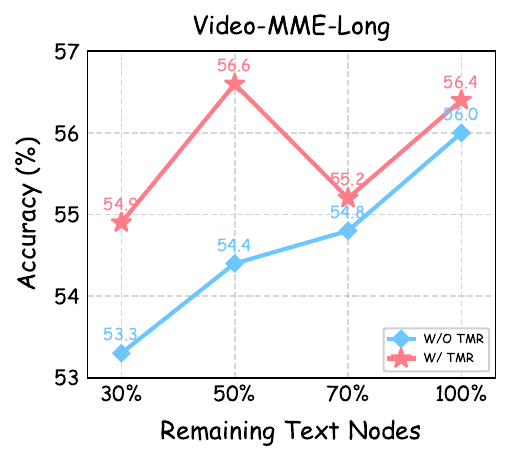}  
  \vspace{-7pt} 
  \caption{Impact of the TMR Mechanism on Video-MME-Long at different compression ratios.}
  \label{Fig9} 
  \vspace{-15pt} 
\end{figure}

As shown in Figure~\ref{Fig9}, consistent with the trends observed in M3-Bench-robot and M3-Bench-web, the model's performance on the Video-MME-Long dataset significantly decreases when the compression ratio exceeds 50\%. After introducing the TMR mechanism, the performance degradation is effectively mitigated, and even the performance at a 50\% compression ratio surpasses that at 30\%. These results are in strong agreement with the previous experiments, indicating that the TMR mechanism can significantly reduce the accuracy loss caused by memory graph compression and retrieve more critical memories for reasoning.

\subsection{Experiments with Other Compression Ratios}

\begin{table}[t]
\vspace{10pt}
\fontsize{9.25}{13}\selectfont
\centering
\begin{tabular}{ccc|c|c}
\toprule[1.5pt]
\multicolumn{3}{c|}{\textbf{Components}} 
& \textbf{M3-Bench-Robot} 
& \textbf{M3-Bench-Web} \\
\hline
\textbf{A} & \textbf{B} & \textbf{C} 
& Accuracy 
& Accuracy \\
\hline

\rowcolor{gray!15}
-- & -- & -- & 30.3 (100.0\%) & 47.9 (100.0\%) \\

\hline
 &  &  & 26.6 (87.8\%)  & 41.8 (87.3\%)  \\
\checkmark &  &  & 27.0 (89.1\%) & 42.6 (88.9\%) \\
 & \checkmark & \checkmark & 28.7 (94.7\%) & 43.9 (91.6\%) \\
 \checkmark & \checkmark &  & 28.4 (93.7\%) & 42.3 (88.3\%) \\
\checkmark &  & \checkmark & 28.9 (95.4\%) & 44.2 (92.3\%) \\
\hline
\rowcolor{blue!3}
\checkmark & \checkmark & \checkmark & \textbf{30.6 (101.0\%)} & \textbf{44.7 (93.3\%)} \\
\bottomrule[1.5pt]
\end{tabular}

\caption{
Ablation study of the compression components. The first row reports the accuracy of the M3-Agent without compression.
\textbf{A}: EMsampling; 
\textbf{B}: entity importance; 
\textbf{C}: embedding similarity.
}
\vspace{-12pt}
\label{table5}
\end{table}

Previously, to highlight the contribution of each module, we conducted experiments under a 30\% compression ratio, and the corresponding results are reported in Table~\ref{table2} of the paper. Here, we further provide ablation results under a 50\% compression ratio, as shown below in Table~\ref{table5} (A: EMsampling; B: entity importance; C: embedding similarity). As observed, compared with random compression, our method still achieves consistent improvements even without the TMR mechanism (i.e., without TMR).

\section{Time Complexity Analysis}

Our compression framework consists of two main modules: EMsampling and EWpruning. Suppose the memory graph contains \( N \) nodes in total. The EMsampling module includes KMeans clustering and Minmax sampling. The clustering stage has a time complexity of \( O(N \log N) \), while Minmax sampling requires computing pairwise distances between nodes, resulting in a time complexity of \( O(N^2) \). Therefore, the overall complexity of EMsampling is \( O(N^2) \). For the EWpruning module, the main computational cost comes from constructing the similarity matrix and the weighted edge matrix, both of which involve pairwise node interactions, leading to a time complexity of \( O(N^2) \). In summary, the overall time complexity of our method is \( O(N^2) \). However, in practical scenarios, for streaming video memory graphs covering 30--60 minutes, each compression run takes less than 2 seconds, demonstrating high efficiency in real-world settings.

\section{Theoretical Analysis}

Within a semantic cluster $C$, let the original node embeddings be $X = \{x_i\}_{i=1}^N$. EMsampling adopts a minmax rule to iteratively select a subset $S \subseteq X$, where the distance is defined as $d(x, S) = \min_{y \in S} \|x - y\|$. At each step, the algorithm selects the node that is hardest to cover (i.e., the one maximizing $d(x,S)$) and adds it to $S$. This greedy process is equivalent to progressively shrinking the covering radius $\rho = \max_{x \in X} d(x,S) = \max_{x \in X} \min_{y \in S} \|x - y\|$. When $\rho$ is small, every original node $x$ has a nearby representative in $S$ within distance $\rho$. As a result, the compressed nodes uniformly cover the original cluster in embedding space, avoiding semantic holes and theoretically leading to minimal accuracy degradation.

For the EWpruning module, we analyze the fused importance score $r$. Let the total scores before and after compression be $X_1$ and $X_2$. According to the scoring function $r_j = b W'_j + (1-b) S'_j$, nodes with smaller $r_j$ are preferentially removed. Let the number of nodes before and after compression be $N$ and $M$. The score loss satisfies $\Delta X = X_1 - X_2 = \sum_{i=1}^N r_i - \sum_{j=1}^M r_j \approx 0$. Therefore, the information loss introduced by compression is theoretically minimal.

\begin{table*}[t]
\fontsize{9.5}{15}\selectfont
  \centering
  \begin{tabular}{lccc}
    \toprule[1.5pt]
    \textbf{Model / Dataset} & \textbf{Office (95)} & \textbf{Meeting Room (68)} & \textbf{Total (163)} \\
    \midrule
    Mem0 (graph) & 42.1 & 45.6 & 43.6 \\
    +Random & 33.7 & 36.8 & 35.0 \\
    +Clustering & 35.8 & 39.7 & 38.0 \\
    +DART & 34.7 & 42.6 & 38.0 \\
    +TimeChat-Memory & 37.9 & 39.7 & 38.7 \\
    +MemoryLLM & 34.7 & 38.2 & 36.2 \\
    +StreamMeCo (w/o TMR) & 40.0 & 44.1 & 41.7 \\
    \bottomrule[1.5pt]
  \end{tabular}
  \caption{Mem0 (graph): Performance comparison of different compression methods on the Office and Meeting Room subsets of M3-Bench-robot (compression ratio = 30\%).}
  \label{table6}
  \vspace{-12pt}
\end{table*}

\section{Experiments on Other Graph-Based Memory Frameworks}

Our method can be readily adapted to other graph-based Agent Memory frameworks. Specifically, we adapt it to the Mem0 (graph)~\cite{r45} framework, whose memory paradigm includes entity types (e.g., persons, events), embeddings, textual memory content, and timestamps. In this setting, nodes connected to entity types such as ``person'' are treated as connected nodes, and we apply the EWpruning module (since this framework does not provide edge weights, we set them to 1). The remaining nodes are treated as isolated textual nodes.

Currently, M3-Agent is the only graph-based Agent Memory framework capable of handling streaming video inputs, while most other frameworks focus on textual inputs. Therefore, we convert the memory graph generated by M3-Agent into textual memories and feed them into Mem0 (graph) for evaluation. We conduct experiments on the office and meeting-room subsets of the M3-Bench-robot dataset. This subset originates from real-world streaming videos and contains 14 videos and 163 questions, making it closer to real-time monitoring scenarios.

The results (compression ratio = 30\%) are shown in Table~\ref{table6}. Our method remains effective in other graph-based Agent Memory frameworks and achieves the best performance. As long as a graph-based Agent Memory framework defines entities such as persons, our method can be directly applied. Moreover, when adapted to other graph-based Agent Memory frameworks, our approach can be seamlessly transferred to different domains, such as long-horizon dialogue scenarios.

\section{Future Works}

Future work can further extend the ideas proposed in this study. First, this work adopts a unified compression ratio for isolated text nodes and connected text nodes; however, future research could explore differentiated and adaptive compression strategies for different node types, so as to better exploit their structural characteristics and information distributions. Second, future studies may investigate the differential compression of episodic nodes and semantic nodes, where episodic nodes primarily focus on specific events occurring in video segments, while semantic nodes emphasize general and relatively stable knowledge distilled from these segments. Effectively preserving highly discriminative event-level descriptions and critical stable semantic memories during compression is an important direction for further enhancing the expressive power and reasoning support of the memory graph. In addition, we observe a noticeable redundancy among text nodes in the M3-Bench-robot dataset. Future work may systematically study the identification and resolution of redundant or even conflicting memories in the memory graph, as well as potential memory poisoning issues. By selectively removing or correcting unreliable memories, the robustness and reliability of the model in complex streaming video scenarios can be further improved. Finally, reducing the time overhead required for memory graph construction in the early stage is also an important research direction. Future work may explore more efficient memory construction mechanisms to lower the cost of generating memory graphs.

\end{document}